\pdfoutput=1
\documentclass[11pt]{article}
\usepackage[table]{xcolor}

\usepackage[preprint]{acl}

\usepackage{times}
\usepackage{latexsym}

\usepackage[T1]{fontenc}
\usepackage[utf8]{inputenc}

\usepackage{microtype}

\usepackage{inconsolata}

\usepackage{graphicx}

\usepackage{fancyvrb, fvextra}
\usepackage{dsfont}
\usepackage{amsmath}
\usepackage{multirow}

\title{Improved LLM Agents for Financial Document Question Answering}

\author{
 \textbf{Nelvin Tan\textsuperscript{1}},
 \textbf{Zian Seng\textsuperscript{1}},
 \textbf{Liang Zhang\textsuperscript{1}},
 \textbf{Yu-Ching Shih\textsuperscript{1}},
 \textbf{Dong Yang\textsuperscript{1}},
 \textbf{Amol Salunkhe\textsuperscript{1}},
\\
\\
 \textsuperscript{1}American Express
 \\
 \tiny{
   \textbf{Correspondence:} \href{thongcainelvin.tan@aexp.com}{thongcainelvin.tan@aexp.com}
 }
}

\begin{document}

\maketitle

\begin{abstract}
Large language models (LLMs) have shown impressive capabilities on numerous natural language processing tasks. However, LLMs still struggle with numerical question answering for financial documents that include tabular and textual data. Recent works have showed the effectiveness of critic agents (i.e., self-correction) for this task given oracle labels. Building upon this framework, this paper examines the effectiveness of the traditional critic agent when oracle labels are not available, and show, through experiments, that this critic agent's performance deteriorates in this scenario. With this in mind, we present an improved critic agent, along with the calculator agent which outperforms the previous state-of-the-art approach (program-of-thought) and is safer. Furthermore, we investigate how our agents interact with each other, and how this interaction affects their performance.
\end{abstract}

\section{Introduction}

Tabular and textual data are ubiquitous in many financial documents. In this paper, we focus on the numerical reasoning ability of large language models (LLMs) on financial data that includes tabular and textual data -- this is a challenging task since LLMs are known to struggle on numerical reasoning for tabular data \cite{Cao2023}, despite their known natural language capabilities \cite{Tan2026}. More generally, LLMs have a low numerical understanding and processsing ability \cite{Yang2024,Chen2024}. Therefore, there is a need to improve the numerical reasoning ability of LLM approaches for tabular and textual data. Recently, \citet{Fatemi2024} presented a multi-agent framework which used LLMs for numerical reasoning (given tabular and textual data) -- at the core of their approach is the use of the critic agent for criticism. Their approach was shown to significantly outperform the chain-of-thought (CoT) approach for various sizes of llama3 models, providing us with a cost-effective alternative to larger LLMs. While the initial results look promising, we noticed that their results seem to disagree with another recent work \cite{Huang2024}, which argued that intrinsic self-correction, which has prompts that are similar to those of the critic agent, does not improve performance -- their experiments are comprehensive but did not include tabular and textual reasoning. Furthermore, it is not obvious whether the results from textual data generalizes to tabular and textual data since the positional relation of information in tables is different from text \cite{Sui2024}. We address this gap by critically examining the effectiveness of the critic agent and show that it does not outperform the CoT approach. With this in mind, we introduce an improved critic agent and a calculator agent which helps to boost the performance of LLMs.

\paragraph{Problem setup.} Given table(s), text, and a numerical question, the goal is to provide an LLM approach (without any fine tuning) that is able to answer the numerical question with high accuracy. 

\section{Related Work}

We first provide a summary on approaches that have been studied for tabular data alone, and then provide a summary on approaches that have been studied for both tabular and textual data.

\paragraph{Tabular data.} \cite{Sui2024} studied the reasoning capabilities of LLMs on tabular data and empirically showed that LLMs have the basic structural understanding capabilities but are far from perfect. The potential usefulness of LLMs for tabular data motivated other works to use LLMs as part of their approach for tabular reasoning. Two major veins of successful approaches are either to harness external tools like Python and SQL \cite{Liu2023,Zhang2024b,Abhyankar2024}, or to decompose the tables before answering the question \cite{Wang2024,Ji2024}. We refer the interested reader to this survey \cite{Zhang2024a}.

\paragraph{Tabular and textual data.} Compared to tabular data, this area is relatively less explored. TAT-LLM \cite{Zhu2024} fine-tuned a smaller LLM (llama 2) for the purpose of question answering from tabular and textual data. While promising, this approach faces challenges related to the high computational costs and memory requirements associated with fine-tuning LLMs. Next, we look at approaches that require no extra training or fine-tuning of LLMs.

\textbf{(1) Program-of-Thought (PoT):} \cite{Chen2023, Phogat2023} This approach involves prompting the LLM directly for an executable Python code. The code is then run to give the final answer. \textbf{(2) Critic:} \cite{Fatemi2024} A critic agent is introduced to refine the previous Chain-of-Thought (CoT) answer \cite{Wei2022} from the LLM. They showed that using \textit{oracle labels}, which are labels that indicate the correctness of the CoT answer, to guide the critic agent (i.e., only using it when the CoT answer is wrong), their approach is able to outperform CoT. \textbf{(3) Domain Specific Language (DSL):} \cite{Phogat2023} Additional LLM prompts are used extract the calculation(s) from the first CoT answer of the LLM, and then present the calculations as a DSL program. The program is then run to produce the final answer.

The previous state-of-the-art approach for financial document question answering is PoT. However, the PoT approach is not purely intrinsic unlike the critic agent (i.e., it relies on external tools). While PoT is one viable solution to resolve the low numerical processing ability of LLMs, it might be dangerous to the company since it would require the execution of a generated program by the LLM (and having employees verify the generated program leads to latency in the company's workflow) -- e.g., the code \texttt{import os; os.rmdir()} is dangerous. Therefore, there is a need for a safer method to improve the LLMs' numerical processing ability.

\paragraph{Main contributions.} In this paper, we build on the multi-agent framework of \citet{Fatemi2024} to study the effectiveness of the critic agent for numerical reasoning from tabular and textual data, and also to introduce the calculator agent. Our main contributions are as follows:
\begin{itemize}
\item We show experimentally that the critic agent is not able to outperform the CoT approach when oracle labels are not available. Our result agrees with the hypothesis of \citet{Huang2024} and generalizes it to the realm of tabular and textual reasoning.
\item We adapt the ideas of \citet{Li2024} to provide an improved critic agent that outperforms the previous critic-agent approach when oracle labels are not available.
\item We introduce a calculator agent and show that in most cases, it outperforms the previous state-of-the-art approach (i.e., PoT) for question answering from financial documents. We also argue that it is a safer approach compared to PoT.
\end{itemize}

\section{Methodology} \label{sec:methodology}

In this section, we extend the multi-agent framework of \citet{Fatemi2024}, which includes only the analyst and critic agent, to include the improved critic agent and the calculator agent -- a potentially useful agent for numerical reasoning. We describe each agent in detail and explain their interactions.

\subsection{Analyst Agent} \label{sec:analyst_agent}

The role of the analyst agent is the following: (i) To provide CoT answer or Python code to solve the given question, (ii) Acts as an intermediary between the user and the other agents. Specifically, it processes outputs returned from other agents (critic, improved critic, and calculator) before returning the final answer to the user -- visual illustration is provided in Appendix \ref{sec:agent_details} (see Figures \ref{fig:CoT_critic}, \ref{fig:CoT_cal}, and \ref{fig:CoT_critic_cal}).

\subsection{Critic Agent and Improved Critic Agent} \label{sec:critic_agent}

The critic agent provides critique on an answer provided for a given question. The critic agent, together with an analyst agent, operates as follows: (1) One LLM prompt is needed to ask the critic agent to critique the CoT answer provided. (2) The critique, together with the previous CoT answer, is then sent back to the analyst agent via another LLM prompt. (3) The analyst agent processes the input and returns the final answer to the user.

We use the critic-agent prompts provided by \citet[Figure 5]{Fatemi2024}. Note that the steps above are similar to the 3-step prompting strategy for self-correction introduced by \citet{Huang2024} if all the steps use the same LLM. We show later in our Section \ref{sec:experiments} that this critic agent does not outperform the CoT approach. With this in mind, we introduce an improved critic agent inspired by \citet{Li2024} which hypothesized that for question answering from textual data (but not tabular and textual data together), the LLM has the ability to gauge its own confidence -- this ability was capitalized by \citet{Li2024} to show that LLM does indeed have the intrinsic ability to self-correct. The improved critic agent, together with the analyst agent, operates as follows: (1) One LLM prompt is needed to ask the critic agent to review the CoT answer provided and decide whether to maintain or update its answer. (2) If the critic agent decides to maintain its answer, then we will output that answer as the final answer. (3) If the critic agent decides to update its answer, then we will send both answers to the analyst agent and ask it to check both answers and the question again, before deciding on the final answer and producing it as an output.

\subsection{Calculator Agent} \label{sec:cal_agent}

The calculator agent takes in a CoT answer and returns the correct answer to all the calculations present in the CoT answer. The calculator agent, together with an analyst agent, operates as follows: (1) One LLM prompt is needed to ask the calculator agent to extract out all the equations in the input (i.e., the previous CoT answer). We do this by few-shot learning, i.e., we give the LLM a few examples of the desired outcome. (2) Python is then used by the calculator agent to evaluate the extracted equations correctly. Our extracted equations only contain numbers, and the symbols `+', `-', `*', `/', `(', and `)'. (3) The correct calculations are then sent back in another LLM prompt, together with the previous CoT answer, to the analyst agent to get the final answer.

This is similar to the DSL approach of \citet{Phogat2023}, but unlike their approach, we directly use Python to evaluate the extracted calculations without producing a DSL program -- our extracted calculations are syntactically simpler, relying only on numbers and a few symbols. Furthermore, \citet{Phogat2023} used a zero-shot approach to extract the calculations, whereas we use a few-shot approach. \citet{Phogat2023} showed that their DSL approach does not outperform PoT, whereas our calculator agent outperforms PoT in most cases (shown later in Section \ref{sec:main_results}).

Regarding step 2, while we have used Python to evaluate the extracted calculations, we could always swap it out for some calculator tool that only performs calculations and nothing else -- this prevents the risk of dangerous code being executed.

\subsection{Interaction between all Agents} \label{sec:all_agents}

The 3 agents stated previously will interact with each other in this manner: (1) The user sends the context (instructions, table, and text) and question over to the analyst agent to receive a CoT answer. (2) The answer is then sent to the critic (or improved critic) agent, and the critic (or improved critic) agent and analyst agent interacts to produce a refined answer -- see Section \ref{sec:critic_agent} for specific steps of the interaction process. (3) The refined answer is then sent to the calculator agent, and the calculator agent and analyst agent interacts to produce a more precise answer -- see Section \ref{sec:cal_agent} for specific steps of the interaction process.

\section{Experiment Details} \label{sec:experiments}

We use a weak LLM (llama3-70B), and a strong LLM (GPT4-turbo), with temperature set to $0$ so that we can achieve consistent results.
% \footnote{For our company's confidentiality purposes, we anonymize our LLM names}.
We use 2 popular financial tabular and textual datasets: TATQA \cite{Zhu2021} and FinQA \cite{Chen2021}. We use \textit{accuracy} is our evaluation metric. Additonal experiment details are provided in Appendix \ref{sec:additional_exp_details}.

\section{Main Results} \label{sec:main_results}

\begin{table*}
    \centering
    \begin{tabular}{ccccccc}
      \hline
      \multirow{2}{*}{\textbf{Approach}} & \multicolumn{2}{c}{\textbf{TATQA}} & \multicolumn{2}{c}{\textbf{FinQA}} & \multicolumn{2}{c}{\textbf{Combined}} \\
       & \textbf{llama3} &  \textbf{GPT4t} & \textbf{llama3} &  \textbf{GPT4t} & \textbf{llama3} &  \textbf{GPT4t} \\
      \hline
      CoT & 72.8\% & 84.4\% & 63.8\% & 74.0\% & 68.3\% & 79.2\% \\
      PoT & 81.2\% & \cellcolor{blue!10}92.1\% & \cellcolor{green!10}70.2\% & 71.6\% & 75.7\% & 81.9\% \\
      CoT+critic & 71.3\% & 84.3\% & 64.1\% & 72.4\% & 67.7\% & 78.4\% \\
      CoT+i-critic & 72.8\% & 85.8\% & 65.4\% & 72.7\% & 69.1\% & 79.3\% \\
      CoT+cal & \cellcolor{blue!10}83.4\% & 90.2\% & \cellcolor{blue!10}72.0\% & \cellcolor{blue!10}75.0\% & \cellcolor{blue!10}77.7\% & \cellcolor{green!10}82.6\% \\
      CoT+critic+cal & 79.7\% & 84.0\% & 67.6\% & 71.3\% & 73.7\% & 77.7\% \\
      CoT+i-critic+cal & \cellcolor{green!10}81.3\% & \cellcolor{green!10}91.1\% & \cellcolor{green!10}70.2\% & \cellcolor{green!10}74.3\% & \cellcolor{green!10}75.8\% & \cellcolor{blue!10}82.7\% \\
      \hline
    \end{tabular}
    \caption{Accuracy of different approaches for TATQA and FinQA. The best result for each column is highlighted in blue and the second best is highlighted in green. The combined accuracy is obtained by taking the average over the two datasets -- the best result for both LLMs involve the calculator agent.}
    \label{tab:TATQA_FinQA_accuracy}
\end{table*}

% \begin{table}
%   \centering
%   \begin{tabular}{ c c c }
%     \hline
%     \textbf{Approach} & \textbf{llama3-70B} & \textbf{GPT4-turbo} \\
%     \hline
%     CoT & 72.8\% & 84.4\% \\
%     PoT & 81.2\% & 92.1\% \\
%     CoT+critic & 71.3\% & 84.3\% \\
%     CoT+i-critic & 72.8\% & 85.8\% \\
%     CoT+cal & 83.4\% & 90.2\% \\
%     CoT+critic+cal & 79.7\% & 84.0\% \\
%     CoT+i-critic+cal & 81.3\% & 91.1\% \\
%     \hline
%   \end{tabular}
%   \caption{Accuracy of different approaches for TATQA.}
%   \label{tab:TATQA_accuracy}
% \end{table}

% \begin{table}
%   \centering
%   \begin{tabular}{ c c c}
%     \hline
%     \textbf{Approach} & \textbf{llama3-70B} & \textbf{GPT4-turbo} \\
%     \hline
%     CoT & 63.8\% & 74.0\% \\
%     PoT & 70.2\% & 71.6\% \\
%     CoT+critic & 64.1\% & 72.4\% \\
%     CoT+i-critic & 65.4\% & 72.7\% \\
%     CoT+cal & 72.0\% & 75.0\% \\
%     CoT+critic+cal & 67.6\% & 71.3\% \\
%     CoT+i-critic+cal & 70.2\% & 74.3\% \\
%     \hline
%   \end{tabular}
%   \caption{Accuracy of different approaches for FinQA.}
%   \label{tab:FinQA_accuracy}
% \end{table}

We have 2 baseline methods: (i) using the analyst agent to produce a CoT answer, and (ii) using the analyst agent to produce an executable Python code which can be executed to give us the final answer (i.e., PoT). For both datasets, we experimented with the following approaches: \textbf{(1) `CoT', `PoT'.} This involves either asking the analyst agent to output the CoT answer (`CoT') or Python code (`PoT') (see Section \ref{sec:analyst_agent} for details). The prompts used can be found in Appendices \ref{sec:LLM_prompt_cot_eg} and \ref{sec:LLM_prompt_pot_eg}. \textbf{(2) `CoT + critic', `CoT + i-critic'.} For `CoT + critic', this involves using the analyst agent and the critic agent to get the final answer (see Section \ref{sec:critic_agent} for details).  For `CoT + critic', this involves using the analyst agent and the improved critic agent to get the final answer (see Section \ref{sec:critic_agent} for details). The prompts used for `CoT + critic' can be found in Appendix \ref{sec:LLM_prompt_cot_critic_eg}, and the prompts used for `CoT + i-critic' can be found in Appendix \ref{sec:LLM_prompt_cot_icritic}. \textbf{(3) `CoT + cal'.} This involves using the analyst agent and the calculator agent to get the final answer (see Section \ref{sec:cal_agent} for details). The prompts used can be found in Appendix \ref{sec:LLM_prompt_eg_CoT_cal}. \textbf{(4) `CoT + critic + cal'.} This involves using the analyst agent, the critic agent, and the calculator agent to get the final answer (see Section \ref{sec:all_agents} for details). The prompts used can be found in Appendix \ref{sec:LLM_prompt_cot_critic_cal_eg}. Before analyzing our results, we state the previous result of \citet{Fatemi2024}: They showed that using llama3-70B, CoT + critic performed $5.19\%$ better than CoT for TATQA, and CoT + critic performed $3.83\%$ better than CoT for FinQA. Our results for TATQA and FinQA are stated in Table \ref{tab:TATQA_FinQA_accuracy}, where llama3-70B and GPT4-turbo are shorten to llama3 and GPT4t, respectively. For TATQA, we have the following observations:
\begin{itemize}
    \item For llama3-70B, CoT + cal is the best performer. Interestingly, CoT + critic performs worse than CoT implying that the critic agent is not useful -- intuitively this can be viewed as the agent overthinking and we provide an example of this in Appendix \ref{sec:LLM_prompt_cot_critic_eg} where the critic agent makes a suggestion to change the previously correct answer from CoT.
    \item For GPT4-turbo, PoT is the best perform. As expected, CoT + critic performs worse than CoT while CoT + i-critic performs better than both CoT +critic and CoT.
\end{itemize}
For FinQA, we have the following observations:
\begin{itemize}
    \item For llama3-70B, CoT + cal is again the best performer. Here, CoT + critic performs slightly better than CoT implying some usefulness of the critic agent. Despite the minor usefulness of the critic agent, CoT + critic + cal does not outperform CoT + cal.
    \item For GPT4-turbo, CoT + cal performs the best, and CoT + i-crtic outperforms CoT + critic. Interestingly, CoT outperforms most of the other approaches except CoT + cal and CoT + i-critic + cal.
\end{itemize}
On average (see `Combined' column in Table \ref{tab:TATQA_FinQA_accuracy}), CoT + cal performs the best when using llama3-70B and CoT + i-critic + cal performs the best when using GPT4-turbo. This suggests that the calculator agent is very useful for numerical tabular and textual reasoning, while the critic agent is not particularly useful. Furthermore, on average, CoT + cal outperforms PoT, which is the previous state-of-the-art approach for financial document question answering.

\paragraph{Remark.} Despite using the same LLM model (llama3-70B) and datasets as \citet{Fatemi2024}, our experiment results differ significantly. This is because when comparing CoT + critic with CoT, \citet{Fatemi2024} only ran CoT + critic on the questions that CoT got wrong (this is stated in the last paragraph of Section 4.1 in their paper) -- their approach has access to oracle labels, which helps in deciding whether to use the critic agent. This is different from the way we evaluate the different approaches, where we run every single approach on the entire dataset. Further comparisons of the critic agents are provided in Appendix \ref{sec:critic_agent_analysis}, and comparison of performance on integer vs.~float answers are provided in Appendix \ref{sec:extra_results}.

\section{Conclusion}

We provide empirical validation that the traditional critic agent does not outperform CoT when oracle labels are not available. With improvement in mind, we present an improved critic agent and a calculator agent, and show empirically that they can improve the LLM's ability in financial question answering safely.

\newpage

\section{Limitations}

In this work, we considered two large language models llama3-70B and GPT4-turbo -- a weaker LLM and a stronger LLM. It is important for us to choose the same model (i.e., llama3-70B) that was used in the previous work \cite{Fatemi2024} which showed the superiority of the critic agent. We acknowledge that we did not exhaustively evaluate a large selection of large language models, but believe that our choices should be sufficient.

Regarding the choice of dataset, we acknowledge that both our datasets comes from the financial domain, which is the scope of this paper. While the domain scope of our dataset might be slightly narrow, we believe that our conclusions should apply to other types of dataset since our key focus was on the ability of our approaches to answer numerical questions with regards to tabular and textual data; our agents do not leverage on any aspects of the financial domain (e.g., we did not use a financial expert agent, or exploit any financial knowledge in our framework).

% Don't show this in arXiv version as well.
% \section{Acknowledgments}
% While this work was conducted at American Express, the conclusions and views expressed are those of the authors and do not necessarily reflect the views of the company.

\bibliography{custom}

\appendix

% \newpage

% \onecolumn

\section{More Details on Agents} \label{sec:agent_details}

\begin{figure}[h]
\centering
\includegraphics[width=0.5\columnwidth]{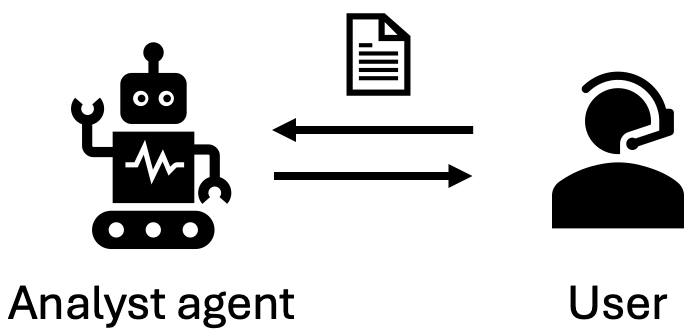}
\caption{Visualization of CoT and coder approach via the analyst agent. Flow: User feeds the table, text, and question to the analyst agent which processes the input (using CoT or PoT) and returns the final answer.}
\label{fig:CoT_or_coder}
\end{figure}

\begin{figure}[h]
\centering
\includegraphics[width=0.9\columnwidth]{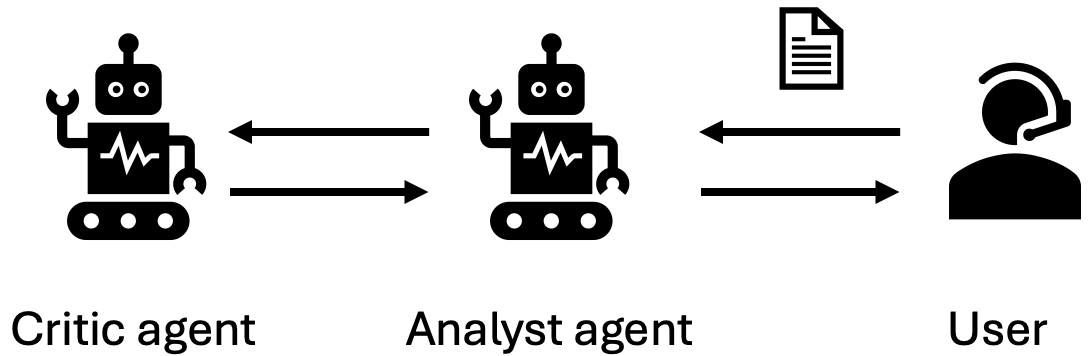}
\caption{Visualization of the critic and analyst agents. Flow: User feeds the table, text, and question to the analyst agent which processes the input, and produces a CoT answer that is passed to the critic agent. The critic agent produces a critique and passes it to the analyst agent, which processes the CoT answer and the critique before returning the final answer.}
\label{fig:CoT_critic}
\end{figure}

\begin{figure}[h]
\centering
\includegraphics[width=0.9\columnwidth]{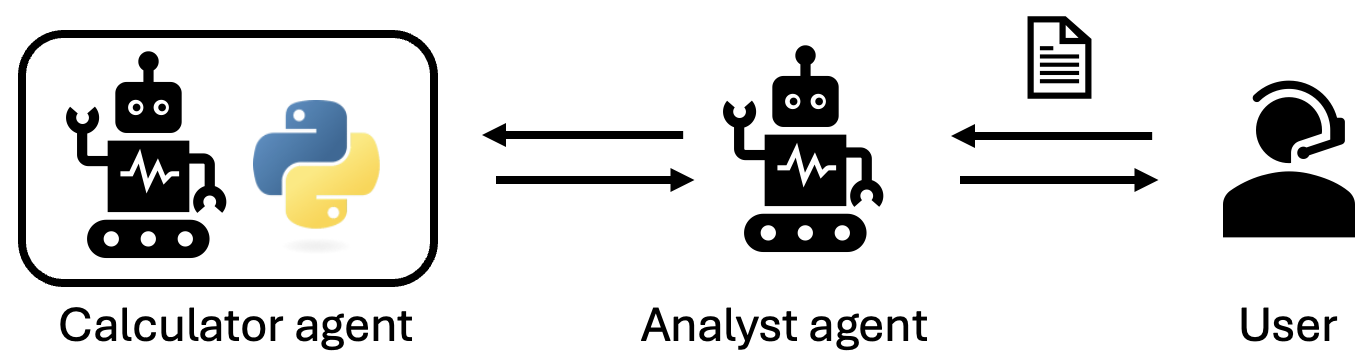}
\caption{Visualization of the calculator and analyst agents. Flow: User feeds the table, text, and question to the analyst agent which processes the input, and produces a CoT answer that is passed to the calculator agent. The calculator agent produces an answer and passes it to the analyst agent, which processes the CoT answer and the calculator agent's answer before returning the final answer.}
\label{fig:CoT_cal}
\end{figure}

\begin{figure}[h]
\centering
\includegraphics[width=0.9\columnwidth]{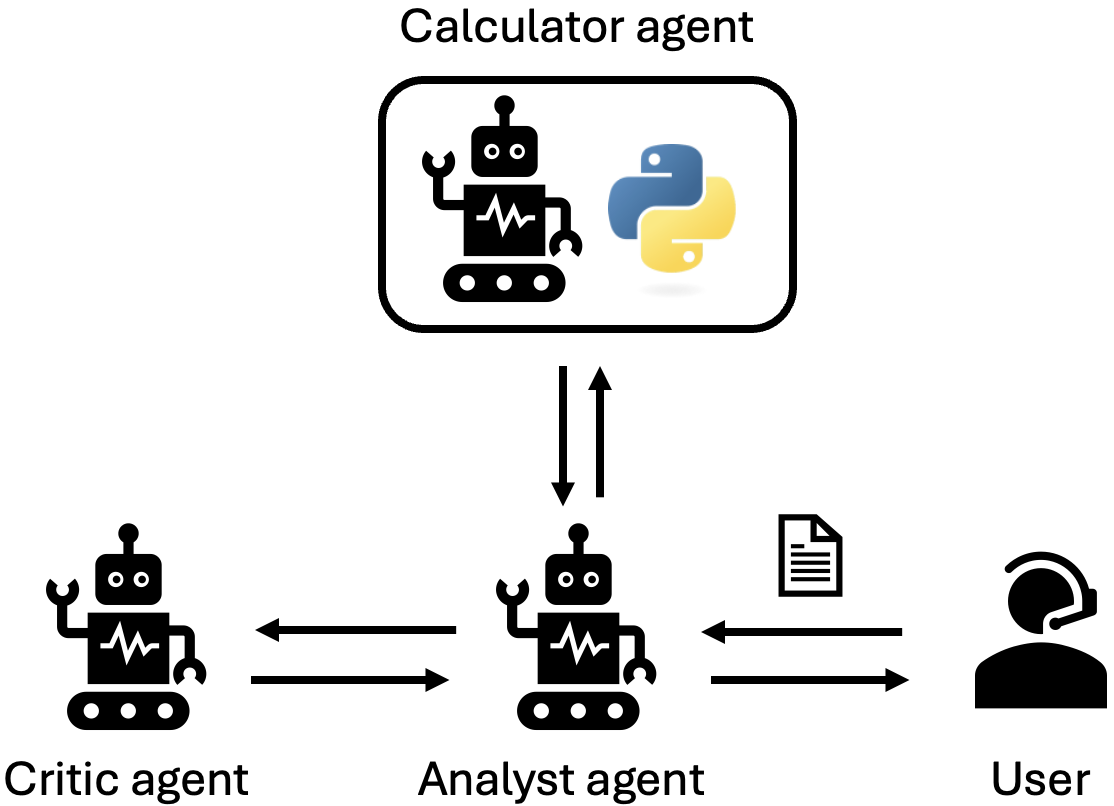}
\caption{Visualization of the critic, calculator, and analyst agents. Flow: User feeds the table, text, and question to the analyst agent which processes the input, and produces a CoT answer that is passed to the critic agent. The critic agent and analyst agent interacts (exactly like in Figure \ref{fig:CoT_critic}) to produce a refined answer. The refined answer is then sent to the calculator agent, and the calculator agent and analyst agent interacts (exactly like in Figure \ref{fig:CoT_cal}) to produce a more precise answer -- which will be the final answer sent to the user.}
\label{fig:CoT_critic_cal}
\end{figure}

A visualization of all the agents are provided in Figures \ref{fig:CoT_or_coder}, \ref{fig:CoT_critic}, \ref{fig:CoT_cal}, and \ref{fig:CoT_critic_cal}.

\section{Additional Experiment Details} \label{sec:additional_exp_details}

We point out that the different agents (introduced in Section \ref{sec:methodology}) are built by prompting one LLM to carry out different tasks -- the same LLM is used for the entire approach. In other words, although we are making multiple calls to the same LLM, we apply the design abstraction of using multiple agents.

\paragraph{Dataset preparation.} We apply our approach to the following popular financial tabular and textual datasets:
\begin{itemize}
    \item  \textbf{TATQA} \cite{Zhu2021}: We use the dev dataset since it has answers to all the questions, along with the type of question (i.e., numerical or non-numerical). We filter out the numerical questions from the dataset, which is a total of 717 questions.
    \item \textbf{FinQA} \cite{Chen2021}: All questions are numerical. We use the dev set since it contains all the answers to the questions -- a total of 883 questions.
\end{itemize}
Questions where any of our approaches exceed the context length are omitted.

\paragraph{Evaluation Metric.} We denote the true answer as $a$ and the estimated answer as $\hat{a}$. For the evaluation metric, we use accuracy defined as
\begin{align*}
    100\%\times\frac{1}{n}\sum_{i=1}^n\mathds{1}\{a=\hat{a}\}, 
\end{align*}
where $n$ is the total number of questions answered, and $\mathds{1}\{\cdot\}$ is the indicator function where $\mathds{1}\{a=\hat{a}\}=1$ if $a=\hat{a}$ and $0$ otherwise. We consider the estimated answer $\hat{a}$ to be equivalent to the true answer $a$ if $\hat{a}$ and $a$ are exactly the same, or if $\hat{a}$ can be rounded to obtain $a$. For example, if $a=0.98$, then $\hat{a}=0.98$ and $\hat{a}=0.979$ are both accepted. 

\section{Comparison of Critic Agents} \label{sec:critic_agent_analysis}

\begin{figure}[t]
  \centering
  \includegraphics[width=1.0\linewidth]{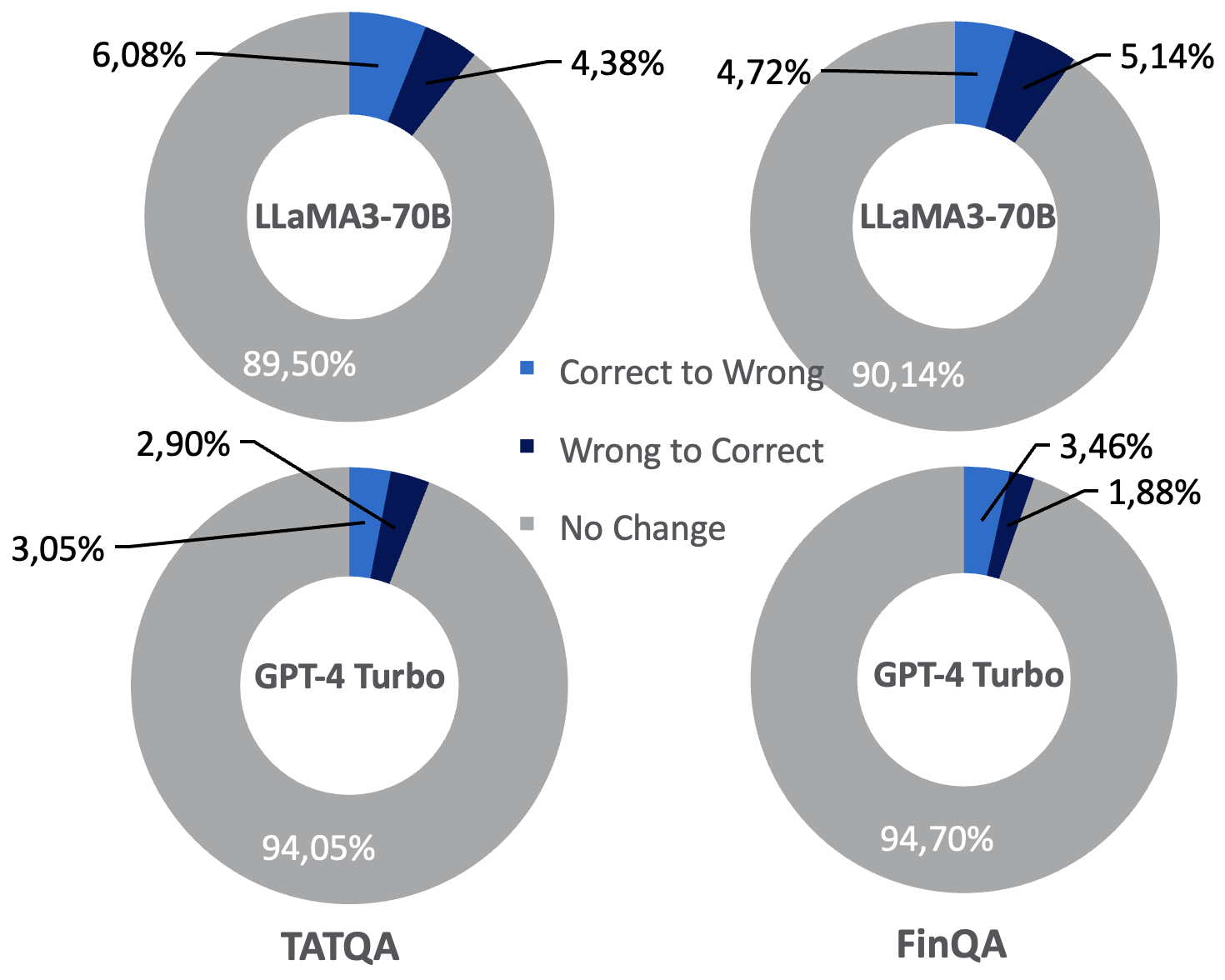}
  \caption {Analysis of the changes in the correctness of answers by the critic agent. Pie charts on the left are for TATQA and pie charts on the right are for FinQA.}
  \label{fig:critic_change}
\end{figure}

\paragraph{Analysis of the critic agent.} We investigate the tendency of the critic agent to switch its answer. The results are displayed in Figure \ref{fig:critic_change}. For both datasets and LLMs, the proportion of answers that were changed from correct to wrong, and the proportion of answers that were changed from wrong to correct are roughly the same. This implies that the critic agent does not have a clear ability to improve the answer.

\begin{figure}[t]
    \centering
    \includegraphics[width=1.0\linewidth]{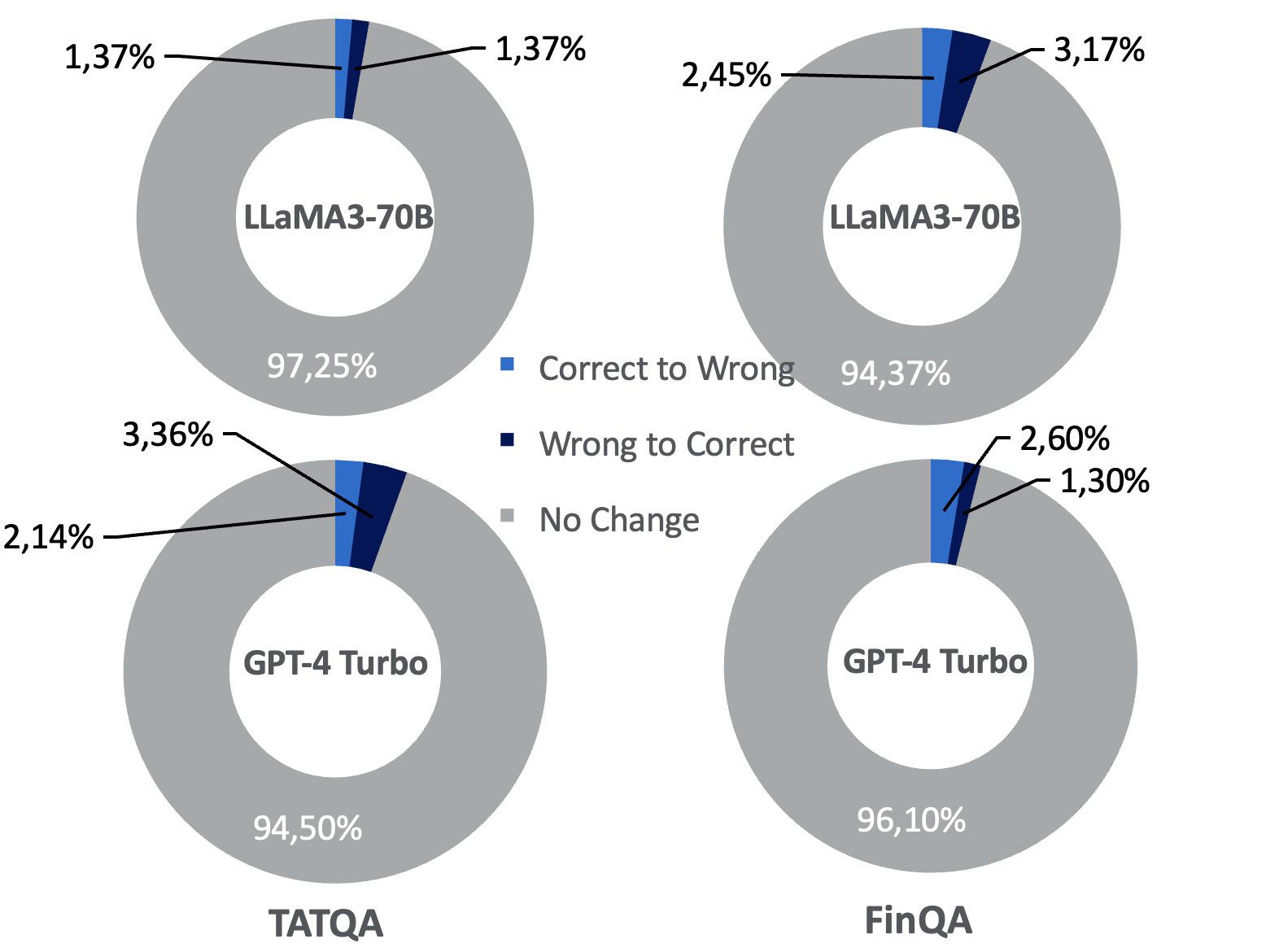}
    \caption {Analysis of the changes in the correctness of answers by the i-critic agent. Pie charts on the left are for TATQA and pie charts on the right are for FinQA.}
    \label{fig:i-critic_change}
  \end{figure}

\paragraph{Analysis of the improved critic agent.} We investigate how the improved critic agent performs given its confidence level. The change rates are visualized in Figure \ref{fig:i-critic_change}. For all cases, except FinQA with GPT-4t, the proportion of answers that were changed from correct to wrong is lower than the proportion of answers that were changed from wrong to correct, indicating the potential effectiveness of the i-critic agent.

Further statistics for TATQA are presented in Table \ref{tab:TATQA_conf_rates}, and our statistics for FinQA are presented in Table \ref{tab:FinQA_conf_rates}. For TATQA, the rate of confident is around the same for both llama3-70B and GPT4-turbo. For FinQA, GPT4-turbo tends to be more confident in its answers compared to llama3-70B. The rate of confidence for both models in FinQA is lower compared to TATQA implying that FinQA is the more challenging dataset. For both datasets, the rate of correct given confident is greater than the rate of incorrect given confident, implying LLM has some accurate sense of its confidence. As expected, GPT4-turbo, the stronger model, has a higher rate of getting correct when it is confident.

\paragraph{Oracle vs.~no oracle.} Table \ref{tab:TATQA_FinQA_accuracy_critics} shows the accuracy comparison between critic and i-critic agents that use oracle labels versus those that do not use oracle labels. We see that both critic and i-critic agents perform better when oracle labels are used, which is expected since the oracle labels help to guide the agents on when to intervene. In all cases (except TATQA with GPT-4t), critc agent with oracle labels outperform the i-critic agent with oracle labels. This differs from the no-oracle setting where i-critic outperforms critic in all cases.

\paragraph{Change analysis in self-criticism.} We proceed to dive deeper into the differences between critic and i-critic agents under the no-oracle setting. We investigate the types of changes made by both agents when the correctness of the answer is flipped after the critique step. We classify the changes into two types:
\begin{itemize}
    \item \textbf{Minor change.} When the agent makes a change to its numerical answer but the steps to getting the numerical answer stay exactly the same. This usually results in a minor change in the final numerical answer.
    \item \textbf{Major change.} When the agent changes the previous steps used to get the numerical answer (e.g., different data is pulled from the table for calculations). This usually results in a major change in the final numerical answer.
\end{itemize}
We study these changes when a correctness flip occurs: The answer changes from correct (before criticism) to wrong (after criticism), or wrong (before criticism) to correct (after criticism). The results are displayed in Table \ref{tab:TATQA_FinQA_critic_changes}. We observe that for all cases (with the exception of FinQA, GPT-4t), the wrong to correct flips have a higher percentage of minor changes compared to the correct to wrong flips, which have a higher percentage of major changes. This might hint that critic agents in general are better at making small adjustments to their previous answers to get the correct answer, rather than completely changing their thought process.

\begin{table*}
    \centering
    \begin{tabular}{ccccc}
      \hline
      \multirow{2}{*}{\textbf{Approach}} & \multicolumn{2}{c}{\textbf{TATQA}} & \multicolumn{2}{c}{\textbf{FinQA}} \\
       & \textbf{llama3} &  \textbf{GPT4t} & \textbf{llama3} &  \textbf{GPT4t} \\
      \hline
      CoT & 72.8\% & 84.4\% & 63.8\% & 74.0\% \\
      CoT+critic (no oracle) & 71.3\% & 84.3\% & 64.1\% & 72.4\% \\
      CoT+critic (oracle) & \cellcolor{blue!10}77.6\% & 87.3\% & \cellcolor{blue!10}68.3\% & \cellcolor{blue!10}75.9\% \\
      CoT+i-critic (no oracle) & 72.8\% & 85.8\% & 65.4\% & 72.7\% \\
      CoT+i-critic (oracle) & 75.0\% & \cellcolor{blue!10}87.9\% & 66.8\% & 75.3\% \\
      \hline
    \end{tabular}
    \caption{Accuracy of different critic approaches for TATQA and FinQA, with and without oracle labels. CoT is also present in the table to act as a baseline comparison. Best performer for each column are highlighted in blue.}
    \label{tab:TATQA_FinQA_accuracy_critics}
\end{table*}

\begin{table*}
    \centering
    \begin{tabular}{ccccccccc}
      \hline
      \multirow{2}{*}{\textbf{Answer Flip}} & \multicolumn{4}{c}{\textbf{TATQA}} & \multicolumn{4}{c}{\textbf{FinQA}} \\
       & \multicolumn{2}{c}{\textbf{llama3}} &  \multicolumn{2}{c}{\textbf{GPT4t}} & \multicolumn{2}{c}{\textbf{llama3}} & \multicolumn{2}{c}{\textbf{GPT4t}} \\
       & minor & major & minor & major & minor & major & minor & major \\
      \hline
      critic (C$\rightarrow$W) & 53.8\% & 46.2\% & 44.4\% & 55.6\% & 22.6\% & 77.4\% & 26.1\% & 73.9\% \\
      critic (W$\rightarrow$C) & 81.4\% & 18.6\% & 83.3\% & 16.7\% & 70.6\% & 29.4\% & 61.5\% & 38.5\% \\
      \hline
      i-critic (C$\rightarrow$W) & 33.3\% & 66.7\% & 71.4\% & 28.6\% & 17.6\% & 82.4\% & 50\% & 50\% \\
      i-critic (W$\rightarrow$C) & 88.9\% & 11.1\% & 81.8\% & 18.2\% & 68.2\% & 31.8\% & 44.4\% & 55.6\% \\
      \hline
    \end{tabular}
    \caption{We present the rates for the different types of changes for different critic approaches. C and W are shorthands for correct and wrong, respectively.}
    \label{tab:TATQA_FinQA_critic_changes}
\end{table*}

\begin{table}
    \centering
    \begin{tabular}{ccc}
      \hline
       & \textbf{llama3} & \textbf{GPT4t} \\
      \hline
      $\text{Rate}(\text{corr}|\text{conf})$ & 75.9\% & 87.0\% \\
      $\text{Rate}(\neg\text{corr}|\text{conf})$ & 24.1\% & 13.0\% \\
      $\text{Rate}(\text{corr}|\neg\text{conf})$ & 41.9\% & 72.2\% \\
      $\text{Rate}(\neg\text{corr}|\neg\text{conf})$ & 58.1\% & 27.8\% \\
      $\text{Rate}(\text{conf})$ & 90.9\% & 91.8\% \\
      \hline
    \end{tabular}
    \caption{Confidence rates for TATQA where $\text{Rate}(\text{corr}|\text{conf})$ means rate of correct answer given that the LLM is confident.}
    \label{tab:TATQA_conf_rates}
\end{table}
  
\begin{table}
    \centering
    \begin{tabular}{ccc}
      \hline
       & \textbf{llama3} & \textbf{GPT4t} \\
      \hline
      $\text{Rate}(\text{corr}|\text{conf})$ & 69.8\% & 75.0\% \\
      $\text{Rate}(\neg\text{corr}|\text{conf})$ & 30.2\% & 25.0\% \\
      $\text{Rate}(\text{corr}|\neg\text{conf})$ & 37.9\% & 54.5\% \\
      $\text{Rate}(\neg\text{corr}|\neg\text{conf})$ & 62.1\% & 45.5\% \\
      $\text{Rate}(\text{conf})$ & 77.5\% & 88.9\% \\
      \hline
    \end{tabular}
    \caption{Confidence rates for FinQA.}
    \label{tab:FinQA_conf_rates}
  \end{table}

\section{Integer vs.~Float Answers} \label{sec:extra_results}

Here we look at how our approaches differ in performance for questions with integer answer versus questions with float answer. The results are shown in Tables \ref{tab:TATQA_int_v_float} and \ref{tab:FinQA_int_v_float}. The key observations are as follows:
\begin{itemize}
    \item For TATQA, with llama3-70B, CoT + cal performs the best for integer answers and PoT performs the best for float answers. With GPT4-turbo, CoT + i-critic performs the best for integer answers and PoT performs the best for float answers.
    \item For FinQA, with llama3-70B, CoT + cal performs the best for integer answers and float answers. With GPT4-turbo, CoT + i-critic + cal performs the best for integer answers and CoT + cal performs the best for float answers.
\end{itemize}
The results justify the importance of the improved critic agent for questions with integer answers.

\begin{table*}
  \centering
  \begin{tabular}{ccccc}
    \hline
    \multirow{2}{*}{\textbf{Approach}} & \multicolumn{2}{c}{\textbf{llama3}} & \multicolumn{2}{c}{\textbf{GPT4t}} \\
     & \textbf{int} &  \textbf{float} & \textbf{int} &  \textbf{float} \\
    \hline
    CoT & \cellcolor{green!10}86.9\% & 61.2\% & 91.5\% & 78.6\% \\
    PoT & 82.7\% & \cellcolor{blue!10}79.9\% & 92.5\% & \cellcolor{blue!10}91.7\% \\
    CoT + critic & 85.3\% & 59.9\% & 91.2\% & 78.6\% \\
    CoT + i-critic & 86.3\% & 61.8\% & \cellcolor{blue!10}94.2\% & 78.9\% \\
    CoT + cal & \cellcolor{blue!10}87.9\% & \cellcolor{green!10}79.7\% & \cellcolor{green!10}93.2\% & 87.8\% \\
    CoT + critic + cal & 83.0\% & 77.0\% & 86.4\% & 81.9\% \\
    CoT + i-critic + cal & 85.9\% & 77.5\% & \cellcolor{green!10}93.2\% & \cellcolor{green!10}89.4\% \\
    \hline
  \end{tabular}
  \caption{Accuracy of the approaches for the integer-answer questions and the float-answer questions of TATQA. The best result for each column is highlighted in blue and the second best is highlighted in green.}
  \label{tab:TATQA_int_v_float}
\end{table*}

\begin{table*}
  \centering
  \begin{tabular}{ccccc}
    \hline
    \multirow{2}{*}{\textbf{Approach}} & \multicolumn{2}{c}{\textbf{llama3}} & \multicolumn{2}{c}{\textbf{GPT4t}} \\
     & \textbf{int} & \textbf{float} & \textbf{int} & \textbf{float} \\
    \hline
    CoT & 75.1\% & 55.9\% & \cellcolor{green!10}79.8\% & 69.7\% \\
    PoT & 75.9\% & \cellcolor{green!10}66.2\% & 75.1\% & 68.9\% \\
    CoT + critic & 72.4\% & 58.3\% & 78.5\% & 67.9\% \\
    CoT + i-critic & 75.5\% & 58.3\% & 78.8\% & 68.2\% \\
    CoT + cal & \cellcolor{blue!10}78.2\% & \cellcolor{blue!10}67.6\% & 79.5\% & \cellcolor{blue!10}71.7\% \\
    CoT + critic + cal & 75.1\% & 62.4\% & 74.1\% & 69.2\% \\
    CoT + i-critic + cal & \cellcolor{green!10}76.3\% & 65.9\% & \cellcolor{blue!10}80.1\% & \cellcolor{green!10}69.9\% \\
    \hline
  \end{tabular}
  \caption{Accuracy of the approaches for the integer-answer questions and the float-answer questions of FinQA. The best result for each column is highlighted in blue and the second best is highlighted in green.}
  \label{tab:FinQA_int_v_float}
\end{table*}

\subsection{Prompt for CoT} \label{sec:LLM_prompt_cot_eg}

We provide the prompt used for the CoT approach.

\begin{Verbatim}[formatcom=\color{red},breaklines,fontsize=\tiny,frame=single]
Read the following texts and table carefully. Present your answer in the following JSON format:
{
    "steps": ["show the calculation steps"],
    "answer": "final numerical answer"
}
### Text
{text}
### Table
{table}

### Question
{question}
\end{Verbatim}

\subsection{Prompt for PoT} \label{sec:LLM_prompt_pot_eg}

We provide the prompt used for the coder approach.

\begin{Verbatim}[formatcom=\color{red},breaklines,fontsize=\tiny,frame=single]
Do not answer the question. Instead, your task is to write some python code that answers the question when it is executed.
Present your python code in the following format:
``
"python code here"
``
### Text
{text}
### Table
{table}

### Question
{question}
\end{Verbatim}

\subsection{Example of Critic Agent Overthinking}  \label{sec:LLM_prompt_cot_critic_eg}

We present one example of the critic agent overthinking from the TATQA dataset. The user's prompt are coloured in red, and the LLM's output are coloured in black. In the following example, the correct answer is \$8590 which tallies with the CoT answer. However, the critic agent found fault with the CoT answer and presented a new answer which is wrong.

\begin{Verbatim}[formatcom=\color{red},breaklines,fontsize=\tiny,frame=single]
Read the following texts and table carefully. Present your answer in the following JSON format:
{
    "steps": ["show the calculation steps"],
    "answer": "final numerical answer"
}
### Text
Other assets consist of the following (in thousands):
(1) In the first quarter of fiscal 2019, we invested 3.0 million Euro ($3.4 million) in 3D-Micromac AG, a private company in Germany. The investment is included in other assets and is being carried on a cost basis and will be adjusted for impairment if we determine that indicators of impairment exist at any point in time.
### Table
|  | Fiscal year-end |  | 
|  | 2019 | 2018 | 
| Assets related to deferred compensation arrangements (see Note 13) | $35,842 | $37,370 | 
| Deferred tax assets (see Note 16) | 87,011 | 64,858 | 
| Other assets(1) | 18,111 | 9,521 | 
| Total other assets | $140,964 | $111,749 | 

### Question
What was the change in Other assets in 2019 from 2018?
\end{Verbatim}

\begin{Verbatim}[breaklines,fontsize=\tiny,frame=single]
Here is the answer in the requested JSON format:

{
    "steps": [
        "Get the value of Other assets in 2019 from the table: $18,111",
        "Get the value of Other assets in 2018 from the table: $9,521",
        "Calculate the change in Other assets: $18,111 - $9,521 = $8,590"
    ],
    "answer": "$8,590"
}
\end{Verbatim}

\begin{Verbatim}[formatcom=\color{red},breaklines,fontsize=\tiny,frame=single]
Review a given context, question, and the response given by an agent. Then, you must reflect on the analysis and provide a detailed critique. Do not round numerical answers.
### Context and Question
### Text
Other assets consist of the following (in thousands):
(1) In the first quarter of fiscal 2019, we invested 3.0 million Euro ($3.4 million) in 3D-Micromac AG, a private company in Germany. The investment is included in other assets and is being carried on a cost basis and will be adjusted for impairment if we determine that indicators of impairment exist at any point in time.
### Table
|  | Fiscal year-end |  | 
|  | 2019 | 2018 | 
| Assets related to deferred compensation arrangements (see Note 13) | $35,842 | $37,370 | 
| Deferred tax assets (see Note 16) | 87,011 | 64,858 | 
| Other assets(1) | 18,111 | 9,521 | 
| Total other assets | $140,964 | $111,749 | 

### Question
What was the change in Other assets in 2019 from 2018?
### Response
Here is the answer in the requested JSON format:

{
    "steps": [
        "Get the value of Other assets in 2019 from the table: $18,111",
        "Get the value of Other assets in 2018 from the table: $9,521",
        "Calculate the change in Other assets: $18,111 - $9,521 = $8,590"
    ],
    "answer": "$8,590"
}
\end{Verbatim}

\begin{Verbatim}[breaklines,fontsize=\tiny,frame=single]
Here's my critique of the response:

**Accuracy:** The response is partially accurate. The calculation of the change in Other assets is correct, but the values used are incorrect.

**Error Analysis:** The mistake lies in the values used for Other assets in 2019 and 2018. The correct values should be $18,111 (not just the value of "Other assets(1)" which is $3.4 million) and $9,521, respectively. The correct calculation should be:

$140,964 (Total other assets in 2019) - $111,749 (Total other assets in 2018) = $29,215

**Improvement Suggestions:**

1. Read the table carefully: The agent should have noticed that the "Other assets" column is a total of three components, and not just the value of "Other assets(1)".
2. Understand the question: The question asks for the change in "Other assets", not just the change in "Other assets(1)".
3. Perform the correct calculation: The agent should have calculated the change in Total other assets, not just the change in one component of Other assets.

**Revised Response:**

{
    "steps": [
        "Get the value of Total other assets in 2019 from the table: $140,964",
        "Get the value of Total other assets in 2018 from the table: $111,749",
        "Calculate the change in Total other assets: $140,964 - $111,749 = $29,215"
    ],
    "answer": "$29,215"
}
\end{Verbatim}

\begin{Verbatim}[formatcom=\color{red},breaklines,fontsize=\tiny,frame=single]
Given the following critique for a response given by an agent, output your answer to the question below in the following JSON format and nothing else:
{
    "steps": ["show the calculation steps and equations (if there are any)"],
    "answer": "final numerical answer"
}
### Context and Question
### Text
Other assets consist of the following (in thousands):
(1) In the first quarter of fiscal 2019, we invested 3.0 million Euro ($3.4 million) in 3D-Micromac AG, a private company in Germany. The investment is included in other assets and is being carried on a cost basis and will be adjusted for impairment if we determine that indicators of impairment exist at any point in time.
### Table
|  | Fiscal year-end |  | 
|  | 2019 | 2018 | 
| Assets related to deferred compensation arrangements (see Note 13) | $35,842 | $37,370 | 
| Deferred tax assets (see Note 16) | 87,011 | 64,858 | 
| Other assets(1) | 18,111 | 9,521 | 
| Total other assets | $140,964 | $111,749 | 

### Question
What was the change in Other assets in 2019 from 2018?
### Response
Here is the answer in the requested JSON format:

{
    "steps": [
        "Get the value of Other assets in 2019 from the table: $18,111",
        "Get the value of Other assets in 2018 from the table: $9,521",
        "Calculate the change in Other assets: $18,111 - $9,521 = $8,590"
    ],
    "answer": "$8,590"
}
### Critique
Here's my critique of the response:

**Accuracy:** The response is partially accurate. The calculation of the change in Other assets is correct, but the values used are incorrect.

**Error Analysis:** The mistake lies in the values used for Other assets in 2019 and 2018. The correct values should be $18,111 (not just the value of "Other assets(1)" which is $3.4 million) and $9,521, respectively. The correct calculation should be:

$140,964 (Total other assets in 2019) - $111,749 (Total other assets in 2018) = $29,215

**Improvement Suggestions:**

1. Read the table carefully: The agent should have noticed that the "Other assets" column is a total of three components, and not just the value of "Other assets(1)".
2. Understand the question: The question asks for the change in "Other assets", not just the change in "Other assets(1)".
3. Perform the correct calculation: The agent should have calculated the change in Total other assets, not just the change in one component of Other assets.

**Revised Response:**

{
    "steps": [
        "Get the value of Total other assets in 2019 from the table: $140,964",
        "Get the value of Total other assets in 2018 from the table: $111,749",
        "Calculate the change in Total other assets: $140,964 - $111,749 = $29,215"
    ],
    "answer": "$29,215"
}
\end{Verbatim}

\begin{Verbatim}[breaklines,fontsize=\tiny,frame=single]
{
    "steps": [
        "Get the value of Total other assets in 2019 from the table: $140,964",
        "Get the value of Total other assets in 2018 from the table: $111,749",
        "Calculate the change in Total other assets: $140,964 - $111,749 = $29,215"
    ],
    "answer": "$29,215"
}
\end{Verbatim}

\subsection{Prompt for Improved Critic Agent}  \label{sec:LLM_prompt_cot_icritic}

We provide the prompts used for the CoT + i-critic approach. The user's prompts are colored in red, and the LLM's outputs are colored in black.

\begin{Verbatim}[formatcom=\color{red},breaklines,fontsize=\tiny,frame=single]
Read the following texts and table carefully. Present your answer in the following JSON format:
{
    "steps": ["show the calculation steps"],
    "answer": "final numerical answer"
}
### Text
{text}
### Table
{table}
### Question
{question}
\end{Verbatim}

\begin{Verbatim}[breaklines,fontsize=\tiny,frame=single]
{CoT output}
\end{Verbatim}

\begin{Verbatim}[formatcom=\color{red},breaklines,fontsize=\tiny,frame=single]
Review your previous answer to the question below using the texts and table. If you are very confident about your answer, maintain your answer. Otherwise, update your answer. Present your final answer in the following JSON format:
{
    "steps": ["show the calculation steps"],
    "answer": "final numerical answer"
}
### Text
{text}
### Table
{table}
### Question
{question}
### Previous answer
{CoT output}
\end{Verbatim}

\begin{Verbatim}[breaklines,fontsize=\tiny,frame=single]
{i-critic agent output}
\end{Verbatim}

At this stage, if the i-critic agent is confident of its output and wants to maintain the previous CoT answer, we will output that answer as the final answer. However, if the critic agent decides to update its answer, then we will send both answers to the analyst agent and ask it to check both answers and the question again, before deciding on the final answer and producing it as an output -- the corresponding prompts are presented below.

\begin{Verbatim}[formatcom=\color{red},breaklines,fontsize=\tiny,frame=single]
You gave two different answers in previous responses. Check the question and your answers again, and give the best answer. Present your final answer in the following JSON format:
{{
    "steps": ["show the calculation steps"],
    "answer": "final numerical answer"
}}
### First previous answer
{CoT output}
### Second previous answer
{i-critic agent output}
\end{Verbatim}

\begin{Verbatim}[breaklines,fontsize=\tiny,frame=single]
{analyst agent output}
\end{Verbatim}

\subsection{Success Example of CoT + Cal} \label{sec:LLM_prompt_eg_CoT_cal}

We provide an example of the prompts that are used. The user's prompt are coloured in red, and the LLM's output are coloured in black. The following example shows how the calculator agent can help correct the CoT answer.
\begin{Verbatim}[formatcom=\color{red},breaklines,fontsize=\tiny,frame=single]
Read the following texts and table carefully. Present your answer in the following JSON format:
{
    "steps": ["show the calculation steps"],
    "answer": "final numerical answer"
}
### Text
Refranchisings and franchisee development — The following table summarizes the number of restaurants sold to franchisees, the number of restaurants developed by franchisees, and gains recognized in each fiscal year (dollars in thousands):
(1) Amounts in 2019, 2018, and 2017 include additional proceeds of $1.3 million, $1.4 million, and $0.2 million related to the extension of the underlying franchise and lease agreements from the sale of restaurants in prior years.
(2)  Charges are for operating restaurant leases with lease commitments in excess of our sublease rental income.
(3) Amounts in 2018 primarily represent $9.2 million of costs related to franchise remodel incentives, $8.7 million reduction of gains related to the modification of certain 2017 refranchising transactions, $2.3 million of maintenance and repair expenses and $3.7 million of other miscellaneous non-capital charges. Amounts in 2017 represent impairment of $4.6 million and equipment write-offs of $1.4 million related to restaurants closed in connection with the sale of the related markets, maintenance and repair charges, and other miscellaneous non-capital charges.
Franchise acquisitions — In 2019 and 2018 we did not acquire any franchise restaurants. In 2017 we acquired 50 franchise restaurants. Of the 50 restaurants acquired, we took over 31 restaurants as a result of an agreement with an underperforming franchisee who was in violation of franchise and lease agreements with the Company. Under this agreement, the franchisee voluntarily agreed to turn over the restaurants. The acquisition of the additional 19 restaurants in 2017 was the result of a legal action filed in September 2013 against a franchisee, from which legal action we obtained a judgment in January 2017 granting us possession of the restaurants. Of the 50 restaurants acquired in 2017, we closed eight and sold 42 to franchisees.
### Table
|  | 2019 | 2018 | 2017 | 
| Restaurants sold to franchisees | — | 135 | 178 | 
| New restaurants opened by franchisees | 19 | 11 | 18 | 
| Proceeds from the sale of company-operated restaurants: |  |  |  | 
| Cash (1) | $1,280 | $26,486 | $99,591 | 
| Notes receivable | — | 70,461 | — | 
|  | $1,280 | $96,947 | $99,591 | 
|  |  |  |  | 
| Net assets sold (primarily property and equipment) | $— | $(21,329) | $(30,597) | 
| Lease commitment charges (2) | — | — | (11,737) | 
| Goodwill related to the sale of company-operated restaurants | (2) | (4,663) | (10,062) | 
| Other (3) | 88 | (24,791) | (9,161) | 
| Gains on the sale of company-operated restaurants | $1,366 | $46,164 | $38,034 | 
### Question
What is the percentage constitution of cash in the total gains on the sale of company-operated restaurants in 2019?
\end{Verbatim}

\begin{Verbatim}[breaklines,fontsize=\tiny,frame=single]
 {
    "steps": [
        "Get the total gains on the sale of company-operated restaurants in 2019 from the table: $1,366",
        "Get the cash proceeds from the sale of company-operated restaurants in 2019 from the table: $1,280",
        "Calculate the percentage of cash in the total gains: ($1,280 / $1,366) * 100%"
    ],
    "answer": "93.2%"
}
\end{Verbatim}

\begin{Verbatim}[formatcom=\color{red},breaklines,fontsize=\tiny,frame=single]
Given the following list of steps, filter out all the equations and list them out in JSON format below. 
Use only numbers without commas (decimal points are allowed), and the symbols "+", "-", "*", "/", "(", ")".
The answer must only include the JSON format and nothing else. 
### Example
### List of steps
['The number of schemes is not provided in the context. Therefore, we cannot calculate the average defined contribution schemes.']
### Answer
{
"answer": []
}
### Example
### List of steps
['First, we need to find the difference between EBITDA and underlying EBITDA for each year.', 'EBITDA (FY19) = 79,046, underlying EBITDA (FY19) = 85,123, so the difference (FY19) = 85,123 - 79,046 = 6,077 thousand.', 'EBITDA (FY18) = 63,954, underlying EBITDA (FY18) = 62,575, so the difference (FY18) = 63,954 - 62,575 = 1,379 thousand.', 'Next, we need to find the average of these differences:', 'Average difference = (6,077 + 1,379) / 2 = 7,456 / 2 = 3,728 thousand dollars.']
### Answer
{
"answer": ["85123-79046=6077", "63954-62575=1379", "(6077+1379)/2=7456/2=3728"]
}
### Example
### List of steps
['Percentage change = (183,191 - 7,081) / 7,081 * 100 = 2,493.634%', 'Alternatively, we can do ($176,110 ÷ $7,081) × 100"]']
### Answer
{
"answer": ["(183191-7081)/7081*100=2493.634", "(176110/7081)*100"]
}

### List of steps
{
    "steps": [
        "Get the total gains on the sale of company-operated restaurants in 2019 from the table: $1,366",
        "Get the cash proceeds from the sale of company-operated restaurants in 2019 from the table: $1,280",
        "Calculate the percentage of cash in the total gains: ($1,280 / $1,366) * 100%"
    ],
    "answer": "93.2%"
}
### Answer
\end{Verbatim}

\begin{Verbatim}[fontsize=\tiny,frame=single]
{
"answer": ["(1280/1366)*100"]
}
\end{Verbatim}

\begin{Verbatim}[formatcom=\color{red},breaklines,fontsize=\tiny,frame=single]
Given the following list of correct calculations for a response given by an agent, improve on the response and output it in the following JSON format:
{
    "steps": ["show the calculation steps and equations (if there are any)"],
    "answer": "final numerical answer"
}
### Response
{
    "steps": [
        "Get the total gains on the sale of company-operated restaurants in 2019 from the table: $1,366",
        "Get the cash proceeds from the sale of company-operated restaurants in 2019 from the table: $1,280",
        "Calculate the percentage of cash in the total gains: ($1,280 / $1,366) * 100%"
    ],
    "answer": "93.2%"
}
{
"correct calculations": "['(1280/1366)*100=93.70424597364568']"
}
\end{Verbatim}

\begin{Verbatim}[breaklines,fontsize=\tiny,frame=single]
 Here is the improved response in the requested JSON format:

``
{
  "steps": [
    "To find the percentage, divide the numerator (1280) by the denominator (1366) and multiply by 100.",
    "The calculation is: (1280 ÷ 1366) × 100"
  ],
  "answer": "93.70%"
}
``
\end{Verbatim}

\subsection{Prompt for CoT + Critic + Cal} \label{sec:LLM_prompt_cot_critic_cal_eg}

We provide the prompts used for the CoT + critic + cal approach. The user's prompts are colored in red, and the LLM's output are colored in black.

\begin{Verbatim}[formatcom=\color{red},breaklines,fontsize=\tiny,frame=single]
Read the following texts and table carefully. Present your answer in the following JSON format.
Keep numbers in your answer of the same scale as those in the table.
For example, do not change 1,000 thousands to 1,000,000 in your answer.
{
    "steps": ["show the calculation steps and equations (if there are any)"],
    "answer": "final numerical answer"
}
### Text
{text}
### Table
{table}

### Question
{question}
\end{Verbatim}

\begin{Verbatim}[breaklines,fontsize=\tiny,frame=single]
{CoT output}
\end{Verbatim}

\begin{Verbatim}[formatcom=\color{red},breaklines,fontsize=\tiny,frame=single]
Review a given context, question, and the response given by an agent. Then, you must reflect on the analysis and provide a detailed critique. Do not round numerical answers.
### Context and Question
### Text
{text}
### Table
{table}

### Question
{question}
### Response
{CoT output}
\end{Verbatim}

\begin{Verbatim}[breaklines,fontsize=\tiny,frame=single]
{critic agent output}
\end{Verbatim}

\begin{Verbatim}[formatcom=\color{red},breaklines,fontsize=\tiny,frame=single]
Given the following critique for a response given by an agent, output your answer to the question below in the following JSON format and nothing else:
{
    "steps": ["show the calculation steps and equations (if there are any)"],
    "answer": "final numerical answer"
}
### Context and Question
### Text
{text}
### Table
{table}

### Question
{question}
### Response
{CoT output}
### Critique
{critic agent output}
\end{Verbatim}

\begin{Verbatim}[breaklines,fontsize=\tiny,frame=single]
{analyst agent output}
\end{Verbatim}

\begin{Verbatim}[formatcom=\color{red},breaklines,fontsize=\tiny,frame=single]
Given the following list of steps, filter out all the equations and list them out in JSON format below. 
Use only numbers without commas (decimal points are allowed), and the symbols "+", "-", "*", "/", "(", ")".
The answer must only include the JSON format and nothing else. 
### Example
### List of steps
['The number of schemes is not provided in the context. Therefore, we cannot calculate the average defined contribution schemes.']
### Answer
{
"answer": []
}
### Example
### List of steps
['First, we need to find the difference between EBITDA and underlying EBITDA for each year.', 'EBITDA (FY19) = 79,046, underlying EBITDA (FY19) = 85,123, so the difference (FY19) = 85,123 - 79,046 = 6,077 thousand.', 'EBITDA (FY18) = 63,954, underlying EBITDA (FY18) = 62,575, so the difference (FY18) = 63,954 - 62,575 = 1,379 thousand.', 'Next, we need to find the average of these differences:', 'Average difference = (6,077 + 1,379) / 2 = 7,456 / 2 = 3,728 thousand dollars.']
### Answer
{
"answer": ["85123-79046=6077", "63954-62575=1379", "(6077+1379)/2=7456/2=3728"]
}
### Example
### List of steps
['Percentage change = (183,191 - 7,081) / 7,081 * 100 = 2,493.634%', 'Alternatively, we can do ($176,110 ÷ $7,081) × 100"]']
### Answer
{
"answer": ["(183191-7081)/7081*100=2493.634", "(176110/7081)*100"]
}

### List of steps
{analyst agent output}
### Answer
\end{Verbatim}

\begin{Verbatim}[breaklines,fontsize=\tiny,frame=single]
{calculator agent output}
\end{Verbatim}

\begin{Verbatim}[formatcom=\color{red},breaklines,fontsize=\tiny,frame=single]
Given the following list of correct calculations for a response given by an agent, improve on the response and output it in the following JSON format:
{
    "steps": ["show the calculation steps and equations (if there are any)"],
    "answer": "final numerical answer"
}
### Response
{analyst agent output}
{
"correct calculations": "{calculator agent output}"
}
\end{Verbatim}

\begin{Verbatim}[breaklines,fontsize=\tiny,frame=single]
{analyst agent output}
\end{Verbatim}

\subsection{Prompt for CoT + I-critic + Cal}  \label{sec:LLM_prompt_cot_icritic_cal}

We provide the prompts used for the CoT + i-critic + cal approach. The user's prompts are colored in red, and the LLM's outputs are colored in black. We first run the prompts in Appendix \ref{sec:LLM_prompt_cot_icritic} to get the output which we call `\{CoT + i-critic output\}'. Afterwards, we have the following prompts:

\begin{Verbatim}[formatcom=\color{red},breaklines,fontsize=\tiny,frame=single]
Given the following list of steps, filter out all the equations and list them out in JSON format below. 
Use only numbers without commas (decimal points are allowed), and the symbols "+", "-", "*", "/", "(", ")".
The answer must only include the JSON format and nothing else. 
### Example
### List of steps
['The number of schemes is not provided in the context. Therefore, we cannot calculate the average defined contribution schemes.']
### Answer
{
"answer": []
}
### Example
### List of steps
['First, we need to find the difference between EBITDA and underlying EBITDA for each year.', 'EBITDA (FY19) = 79,046, underlying EBITDA (FY19) = 85,123, so the difference (FY19) = 85,123 - 79,046 = 6,077 thousand.', 'EBITDA (FY18) = 63,954, underlying EBITDA (FY18) = 62,575, so the difference (FY18) = 63,954 - 62,575 = 1,379 thousand.', 'Next, we need to find the average of these differences:', 'Average difference = (6,077 + 1,379) / 2 = 7,456 / 2 = 3,728 thousand dollars.']
### Answer
{
"answer": ["85123-79046=6077", "63954-62575=1379", "(6077+1379)/2=7456/2=3728"]
}
### Example
### List of steps
['Percentage change = (183,191 - 7,081) / 7,081 * 100 = 2,493.634%', 'Alternatively, we can do ($176,110 ÷ $7,081) × 100"]']
### Answer
{
"answer": ["(183191-7081)/7081*100=2493.634", "(176110/7081)*100"]
}

### List of steps
{CoT + i-critic output}
### Answer
\end{Verbatim}

\begin{Verbatim}[breaklines,fontsize=\tiny,frame=single]
{calculator agent output}
\end{Verbatim}

\begin{Verbatim}[formatcom=\color{red},breaklines,fontsize=\tiny,frame=single]
Given the following list of correct calculations for a response given by an agent, improve on the response and output it in the following JSON format:
{
    "steps": ["show the calculation steps and equations (if there are any)"],
    "answer": "final numerical answer"
}
### Response
{CoT + i-criitic output}
{
"correct calculations": "{calculator agent output}"
}
\end{Verbatim}

\begin{Verbatim}[breaklines,fontsize=\tiny,frame=single]
{analyst agent output}
\end{Verbatim}

\end{document}